\documentclass[11pt]{article}
\usepackage{xcolor}
\usepackage{graphicx}
\usepackage{fullpage}
\usepackage{amsmath,amssymb}
\usepackage{booktabs}
\usepackage{url}
\usepackage{array}
\newcolumntype{C}[1]{>{\centering\let\newline\\\arraybackslash\hspace{0pt}}m{#1}}
\title{Reinforcement Learning for Multi-Objective Optimization of Online Decisions in High-Dimensional Systems}

\author{Hardik Meisheri$^1$, Vinita Baniwal$^1$, Nazneen N Sultana$^1$, \\
Balaraman Ravindran$^2$, Harshad Khadilkar$^{1,3}$\\ \\
$^1$TCS Research,\\$^2$Robert Bosch Centre for Data Science and AI, IIT Madras, \\$^3$IIT Bombay}
\date{}

\begin{document}

\maketitle

\begin{abstract}
  This paper describes a purely data-driven solution to a class of sequential decision-making problems with a large number of concurrent online decisions, with applications to computing systems and operations research. We assume that while the micro-level behaviour of the system can be broadly captured by analytical expressions or simulation, the macro-level or \textit{emergent} behaviour is complicated by non-linearity, constraints, and stochasticity. If we represent the set of concurrent decisions to be computed as a vector, each element of the vector is assumed to be a continuous variable, and the number of such elements is arbitrarily large and variable from one problem instance to another. We first formulate the decision-making problem as a canonical reinforcement learning (RL) problem, which can be solved using purely data-driven techniques. We modify a standard approach known as advantage actor critic (A2C) to ensure its suitability to the problem at hand, and compare its performance to that of baseline approaches on the specific instance of a multi-product inventory management task. The key modifications include a parallelised formulation of the decision-making task, and a training procedure that explicitly recognises the quantitative relationship between different decisions. We also present experimental results probing the learned policies, and their robustness to variations in the data.
\end{abstract}



\section{Introduction} \label{sec:intro}
Reinforcement Learning (RL) algorithms have been successfully applied to a broad range of sequential decision-making applications, including software-based gameplay \cite{mnih2015human,silver2016mastering} and physical systems such as autonomous driving \cite{shalev2016safe} or flight dynamics \cite{ng2006autonomous}. An essential feature of these applications is the ability to plan strategies that maximize a long term discounted future reward under constraints. This property makes RL a natural candidate for a broader class of problems, where the systems are not necessarily monolithic but can be distributed. 
We analyse the applicability of RL to high-dimensional stochastic systems, and the modifications required for its use in real-world contexts such as industrial operations. We analyse a system with dynamics described by vector differential equations because the sample application is from operations research. However, the same approach should work for any instance where a large number of interdependent decisions needs to be computed without online search, and high accuracy (for example, resource allocation in computing systems or service rates in queueing systems).

\begin{figure}[h]
\centering
\includegraphics[width=0.55\textwidth]{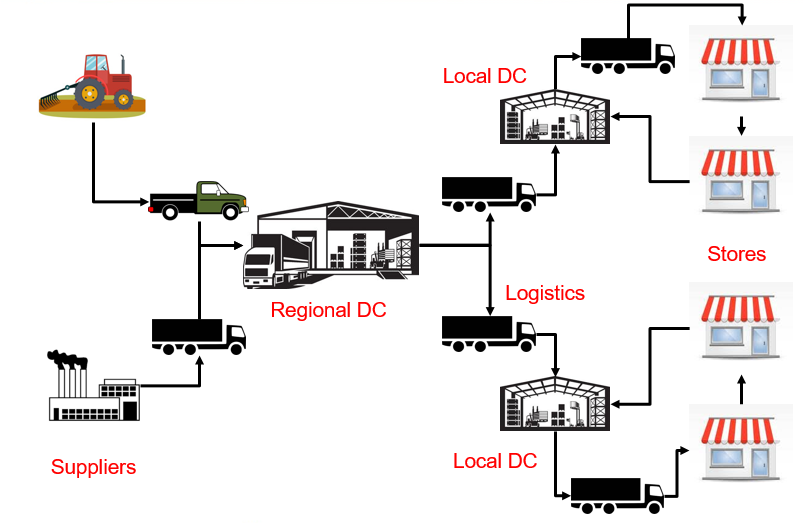}
\caption{Illustration of inventory management in a retail supply chain. The goal of the system is to maintain inventory of a range of products in the stores, while minimising costs. (DC = Distribution center)}
\label{fig:sc-eg}
\end{figure}

We demonstrate the nuances of the generic problem using the multi-product two-echelon retail inventory management scenario illustrated in Fig. \ref{fig:sc-eg}. A moderately large retail business may be composed of approximately 1,000 stores, with each store selling approximately 100,000 product types. The objective of each store is to maintain its inventory at an optimal level, by balancing the depletion process (through product sales and wastage of perishables) with the replenishment process (re-stocking from a warehouse). The replenishment is carried out through periodic deliveries by truck from the warehouse to the store. The products destined for a specific store are typically carried on the same truck (or set of trucks), which imposes constraints on their replenishment quantities. There are multiple tradeoffs involved in this resource allocation task, including maintenance of minimum inventory, minimisation of wastage due to products going past their sell-by dates, and capacity allocation to each product on the truck. Furthermore, the decisions in each time step strongly affect the sequence of inventories for all products in the long term. In this paper, we focus on the last step of the supply chain shown in Fig. \ref{fig:sc-eg}; the replenishment of store inventory from the local distribution centre.

The reason for using reinforcement learning is the complexity of the problem, including non-linearity, non-stationarity, and scale of the state-action space. We survey the existing techniques in detail in Section \ref{sec:related}. The problem is described in detail in Section \ref{sec:desc}, and the proposed parallelised decision-making algorithm is developed in Section \ref{sec:method}. Using data from public sources, we show a comparison between the proposed algorithm and a set of baselines in Section \ref{sec:results}. We show that by using the unique properties of the problem and the ability of RL to learn nuanced control policies, it is possible to achieve high solution quality with a minimally complex and explainable model.

The contribution of this paper is threefold. First, we show that the high-dimensional continuous-time dynamical system decision-making problem can be formulated as a standard reinforcement learning problem. Second, we instantiate the generic formulation using the example of inventory management, a common problem in operations research. Finally, we show how standard reinforcement learning algorithms can be modified to address this class of problem.

\section{Related work} \label{sec:related}

We classify prior literature related to this work into various categories, including the problem domain, traditional approaches for control of such systems, data-driven approaches such as adaptive control, approximate dynamic programming (ADP) and imitation learning (IL), control algorithms for supply chain and inventory management, and the use of reinforcement learning in related problem areas.

\textbf{Problem forms: }
Dynamical systems are typically distinguished based on the form of state that they aim to control. The simpler form is a scalar state such as the rate of a chemical reaction \cite{gustafsson1983dynamic} or temperature of a boiler \cite{liu2006neuro}. A more complex version is the multivariable control problem, involving a vector of state variables which may or may not be directly dependent on each other. The canonical multivariable system with dynamics has a state represented by a vector of variables, and their evolution is modelled using matrix differential equations \cite{ogata2002modern}. Examples of multivariable problems with directly dependent state are the position and its derivatives of an inverted pendulum \cite{pathak2005velocity}, while indirectly dependent states are found in transportation networks \cite{de2010multi} and inventory management in supply chains \cite{van1997production}.

\textbf{Control of dynamical systems: }
The preferred control approach for multivariable problems is to use a linear time invariant (LTI) model of the system and to design a controller using classical methods in the frequency domain or using state feedback \cite{bouabdallah2004pid,golnaraghi2010automatic}. Non-linear versions of the problem are solved using techniques such as sliding mode \cite{utkin2009sliding} and model predictive control \cite{mayne2000constrained}, while robust control under stochastic disturbances is handled using techniques such as $H_\infty$ \cite{doyle1989state}. The key takeaway from these techniques is that they address one complex aspect of the problem in isolation. For example, frequency domain control synthesis addresses stability and robustness when the system dynamics are linear and known, while state feedback control works with high dimensionality under the same assumptions. Sliding mode and model predictive control also assume that the system dynamics are known, and typically are only applicable in low dimensional problems. Robust control techniques allow the noise or disturbance to be arbitrary, but the system dynamics are known. A separate class of techniques addresses the problem of system identification, which focuses on models with unknown dynamics \cite{ljung1998system,bestle1983canonical}. However, these methods do not include control design with this function. In fact, there are results in many classes of problems where a simple closed loop with system identification followed by a traditional controller has no stability guarantees whatsoever \cite{doyle1978guaranteed}. 

\textbf{Data-driven control approaches: }
Given the limitations of a pre-designed control algorithm, we describe data-driven adaptive techniques in prior work. There exists a large volume of existing literature on adaptive control \cite{ioannou1996robust,aastrom2013adaptive}, where the control law is defined as a functional relationship while the parameters are computed using empirical data. However, adaptive control typically requires analytical models of the control and adaptation laws. Approximate Dynamic Programming (ADP) \cite{bertsekas2005dynamic,powell2007approximate} has a similar dependence on analytical forms of the value function, at least as a weighted sum of basis functions for solving the projected Bellman equation. It also requires models of the state transition probabilities and stage costs, which may not be available in the current context, since the noise and system dynamics are assumed to be unknown. The closest form of ADP for problems of the current type is the literature on Adaptive Critics \cite{si2004handbook}, which has considerable overlap with reinforcement learning.

ADP in the policy space solves the problem by using policy gradients to compute the optimal policy parameters, each of which defines a stationary policy. Actor-critic based reinforcement learning could be viewed as an extension of this approach, where the `policy parameters' are actually the parameters of the critic and actor networks and are computed using simulation (or equivalently, \textit{trained}). ADP has been used in prior literature for relatively large task allocation problems in transportation networks \cite{godfrey2002adaptive,topaloglu2006dynamic}. These studies use non-linear approximations of the value function, but the forms are still analytically described. Furthermore, they require at least a one-step rollout of the policy. This may not be feasible in the current context, since the dimensionality is high and each action is continuous (or at least finely quantised), and the goal is not to track some reference signal as in the standard linear quadratic regulator (LQR) \cite{ogata2002modern}.

Imitation learning (IL) is a well-known approach for learning from expert behaviour without having any need of a reward signal and with the simplicity of a supervised learning. This approach assumes that expert decisions have considered all the constraints of the system in order to accomplish the objective. IL has been used in variety of problems including games \cite{Stephane2010}, 3D games \cite{Jack2018}, and robotics \cite{Yan2017}. The inherent problems of design complexity and performance limitations apply here as well; to the definition of the expert policy rather than to the IL algorithm. Additionally, the general form of the problem may not admit an obvious expert policy to train with.

\textbf{Reinforcement learning in related areas: }
The majority of research in reinforcement learning has been for computing actions in games such as Atari and chess. Games are attractive RL applications because they are easy to simulate and to understand. Deep Q-Network \cite{mnih2015human} and its variants \cite{van2016deep,schaul2015prioritized} were used to achieve superhuman performance on Atari using raw pixel inputs. Recently, policy gradient methods have increasingly become the state-of-the-art methods for handling continuous action spaces \cite{islam2017reproducibility}. Deep Deterministic Policy Gradients (DDPG) \cite{ddpg_modified} improves upon the basic advantage actor critic \cite{konda2000actor} by using the actions as inputs to the critic, and using sampled gradients from the critic to update the actor policy. DDPG has been shown to work well on continuous action spaces, although results so far are limited to a few (less than 10) action outputs. Trust Region Policy Optimization \cite{schulman2015trust} and Proximal Policy Optimization \cite{schulman2017proximal} have also proven effective for optimal control using RL, but these are on-policy computationally expensive algorithms and are difficult to apply where episodes are not naturally finite-horizon. Approaches for multiple continuous actions such as Branching DQN \cite{tavakoli2018action} still have significant growth in size with the state space. A reduction in action space dimension by learning representations of the actions \cite{chandak2019learning} is difficult when there are complex constraints on specific subsets of the actions and states (Sec \ref{sec:desc}).

In system dynamics, there is significant work in the computation of torque commands for robotic applications \cite{powell2012ai,kober2013reinforcement} including locomotion \cite{kohl2004policy} and manipulation \cite{theodorou2010reinforcement}. A number of these methods are model-based \cite{nagabandi2018neural}, because of the availability of accurate dynamic models of the robots. The action spaces are naturally continuous and are either discretised for tractability, or are represented by function approximations. Alternatively, the policy is parameterised for simplicity \cite{theodorou2010reinforcement}. The key point of complexity is the curse of dimensionality, which is much more acute in the current context than in typical robotic applications with fewer than 10 degrees of freedom. A recent approach for exploration in large state-action spaces is learning by demonstration \cite{nair2018overcoming}. However, this requires the equivalent of an expert policy for imitation learning.

Intelligent transportation systems \cite{bazzan2013introduction} also require online decisions for managing transportation network operations for maximizing safety, throughput, and efficiency. Adaptive traffic signal control has been a major challenge in transportation systems. In literature, it has been solved by modeling it as a multiple player stochastic game, and solve it with the approach of multi-agent RL \cite{shoham2007if,busoniu2008comprehensive,el2013multiagent}. However, these approaches are difficult to scale. Other approaches \cite{khadilkar2018scalable} tackle the scalability issue by dividing the global decision-making problem into smaller pieces, with both local and global performance affecting the reward. We use a similar approach in this work, as described in Section \ref{sec:method}.

\textbf{Inventory management problems: }
We finally summarise existing literature on supply chain control and inventory management, which is introduced in Section \ref{sec:desc} as a specific instance of the generic control problem (\ref{eq:mainsysdyn})-(\ref{eq:repldyn}). The operations research community has addressed this problem in detail because of its commercial implications \cite{silver1981operations}, and most of the literature considers a two-echelon inventory management problem with varying degree of complexity \cite{nahmias1994optimizing, Nahmias1993, lee1993material}. We summarize the key literature below. One major challenge that prior work fails to address is handling of multiple product decisions at a time, partly due to scaling difficulties with mixed-integer programming or meta-heuristics.

Traditionally, two-echelon retail inventory problems are addressed using linear programming or approximate dynamic programming \cite{neuro_dynamic_approach}. Inventory management instances at relatively small scale are solved as joint assortment-stocking problems using mixed-integer linear programming \cite{smith2000management,caro2010inventory} and related techniques such as branch-and-cut \cite{coelho2014optimal}. However, these techniques are limited to problems with a handful of product types and short time horizons. Furthermore, the computation times for such methods can be significant. When an algorithm needs to be run every few hours, the computation time limits can be of the order of a few minutes for close to one million products. Using exact methods does not appear to be a feasible approach in this context. Implementations at practical business scales typically operate with simple heuristics such as threshold-based policies \cite{condea2012rfid} or formulae based on demand assumptions \cite{cachon1997campbell,silver1979simple}. Adaptive critic \cite{shervais2003intelligent} and reinforcement learning \cite{giannoccaro2002inventory,jiang2009case,mortazavi2015designing} approaches are also reported in literature, but again tend to focus on single-product problems.

In the rest of this paper, we show how a sequential decision-making problem with a large number of concurrent actions can be formulated as a canonical reinforcement learning problem. Using the example of a multi-product inventory management scenario, we show how existing RL algorithms can be modified to compute replenishment decisions online at large scale. 

\section{Problem description} \label{sec:desc}

As we outlined in Section \ref{sec:intro}, the generic decision-making problem that we consider relates to dynamical systems described by vector differential equations along the time dimension. However, since the reinforcement learning algorithm described subsequently is agnostic to the actual system model, one can easily replace the analytical expressions by a discrete-time model or a simulation.

\textbf{Abstract version of the problem: }
We consider a system with continuous-time dynamics where inputs are provided in a long sequence of discrete time intervals. This form captures a large variety of real-world systems with digital controllers. The system dynamics between two time steps are given by,
\begin{equation}
\dot{\bf{x}}(z) = F({\bf{x}}(z)) + {\bf{w}}(z), \label{eq:mainsysdyn}
\end{equation}
where $\bf{x}$ is the vector of $p$ state variables, $F({\bf{x}}): \mathbb{R}^p\rightarrow \mathbb{R}^p$ is a (possibly nonlinear) function, $\bf{w}$ is external noise, and $z$ denotes continuous time. The initial state $\bf{x}(0)$ is specified and arbitrary. The control input is provided at discrete time intervals $t$. Without loss of generality, let us assume that $t\in\mathbb{I}^+$, the set of positive integers. The effect of control is an instantaneous change in the state $\bf{x}$ at time $t$ and is given by,
\begin{equation}
{\bf{x}}(t)^+ = {\bf{x}}(t)^- + B\,{\bf{u}}(t). \label{eq:repldyn}
\end{equation}
The control problem specifies that some predefined reward function of the system states $\bf{x}$, a set of parameters $\bf{\theta}$, and a discount factor $\gamma$, be maximised in the long term. Formally, we state this as the computation of the optimal control vector,
\begin{align}
\bf{u}^*(t) & = \mathrm{arg}\,\mathrm{max} \int_{t}^{\infty} \gamma^{z-t}\,r\left(\hat{\bf{x}}(z),{\bf{u}}(t),\bf{\theta}\right)\,\mathrm{d}z,\label{eq:opt}\\ 
\text{subject to } & {\bf{h}}\left({\bf{x}}(t),{\bf{u}}(t)\right)  \leq 0, \;\; {\bf{g}}\left({\bf{x}}(t),{\bf{u}}(t)\right) = 0, \nonumber
\end{align}
where $\hat{{\bf{x}}}(z)$ is an estimate of the future state trajectory, $\bf{h}$ is a set of inequality constraints on the values of the state and control inputs, and $\bf{g}$ is a set of equality constraints. We assume that system dynamics $F$ are unknown but the control input matrix $B$ is known, and both $F$ and $B$ could in general be time dependent. However, the variation in nature of $F$, $B$, and $\bf{w}$ is slow enough to build reasonable estimators $\hat{F}$ and $\hat{{\bf{w}}}$, thus admitting an explicit or implicit predictor for the trajectory $\hat{\bf{x}}$ given historical values of $\bf{x}$. The noise statistics of $\bf{w}$ can be arbitrary.

\textbf{Reinforcement learning formulation: }
The general dynamics of system evolution and control input are given by (\ref{eq:mainsysdyn}) and (\ref{eq:repldyn}), and the optimal control problem is defined by (\ref{eq:opt}). We now show how these relations can be framed as a standard RL problem. Subsequently, we instantiate the generic framework using an inventory management scenario.

The problem can be modeled as a Markov Decision Process ($\mathcal{S}, \mathcal{A}, \mathcal{T}, \mathcal{R}, \gamma$), where $\mathcal{S}$ represents the state space defined by a feature map $\bf{f}({x}):\mathbb{R}^p \rightarrow \mathbb{R}^{m\times p}$, $\mathcal{A}$ denotes the decision or action space ${\bf{u}}(t)\in \mathbb{R}^p$, $\mathcal{T}$ represents transition probabilities from one combination of state and action to the next, $\mathcal{R}$ denotes the rewards, and $\gamma$ is the discount factor for future rewards. Given the optimisation task as defined in (\ref{eq:opt}), the objective is equivalent to a discounted sum of aggregated discrete rewards $R(n)$,
\begin{align}
\text{Max: }& \int_{t}^{\infty} \gamma^{z-t}\,r\left(\hat{\bf{x}}(z),{\bf{u}}(t),{\bf{\theta}}\right)\,dz \nonumber \\
 \equiv \text{Max: } & \sum_{n=t}^{\infty}\,\left(\gamma^{n-t}\,\int_{n}^{n+1} \gamma^{z-n}\,r\left(\hat{\bf{x}}(z),{\bf{u}}(t),{\bf{\theta}}\right)\,\mathrm{d}z\right) \nonumber \\
& \equiv \text{Max: }\sum_{n=t}^{\infty}\,\gamma^{n-t}\,R(n). \label{eq:reward}
\end{align}
While (\ref{eq:opt}) is a standard optimisation problem, the arbitrary nature of $F$, $B$, and $\bf{w}$, the possible nonlinearity of $\bf{h}$ and $\bf{g}$, and the online response requirement of computing ${\bf{u}}(t)$ makes it a difficult proposition to solve using traditional approaches. Even with simple linear forms of $F$, large dimensionality of $\bf{x}$ and $\bf{u}$ can make the problem intractable. On the other hand, the reward structure (\ref{eq:reward}) readily admits the use of RL for computing the inputs ${\bf{u}}(t)$. Model-free RL can implicitly model the estimator for noise as well as the future state trajectory in order to maximise the discounted long-term reward, requiring only (i) a feature set $\bf{f}({x})$ as the input, and (ii) a mechanism (such as simulation) for rolling out the effect of actions on the states in a closed-loop fashion.

\begin{figure}[h]
    \centering
    \includegraphics[width=0.65\textwidth]{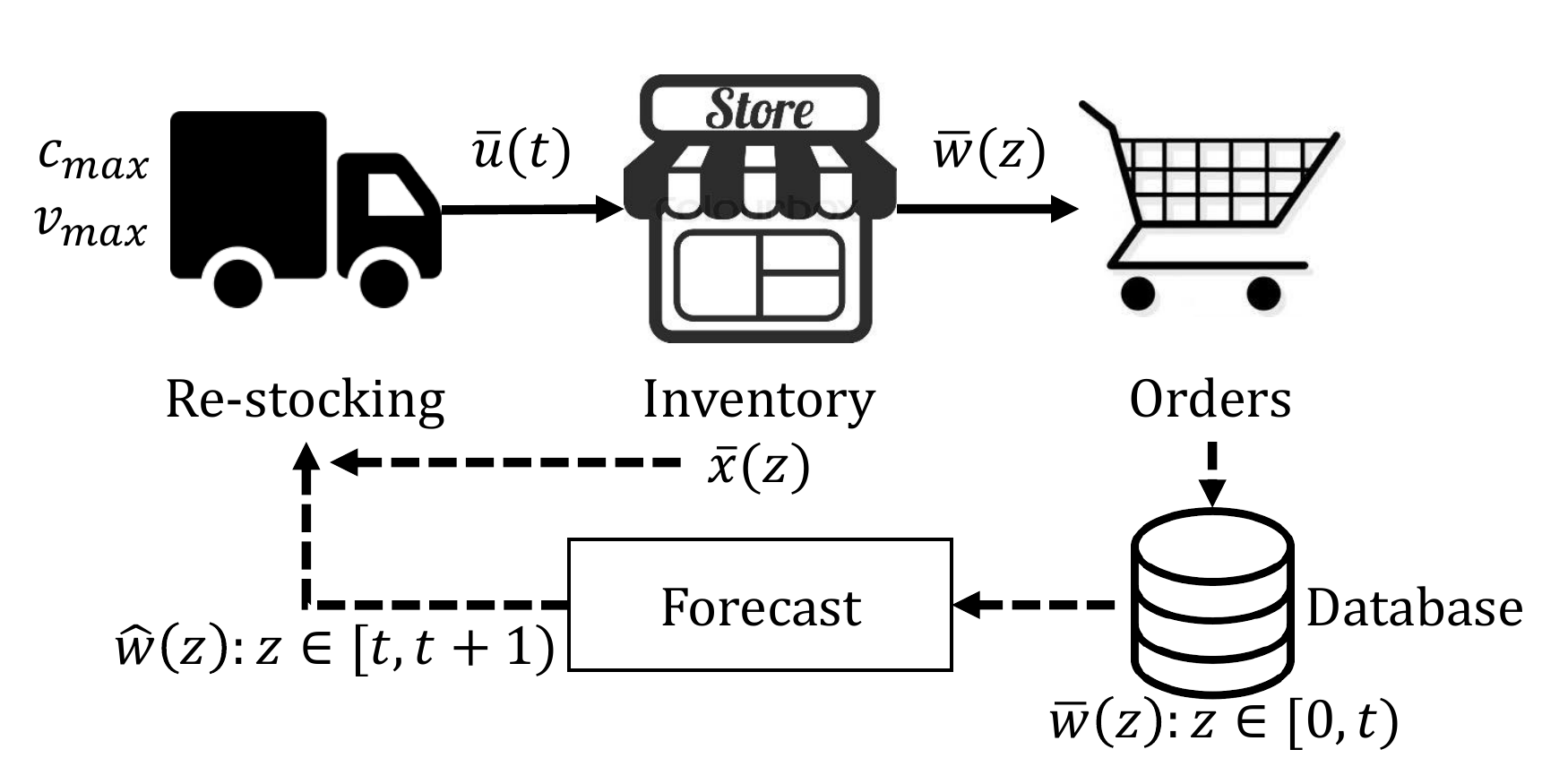}
    \caption{Architecture diagram of inventory management problem.}
    \label{fig:sysarch}
\end{figure}

\textbf{Multi-product inventory management: }
We instantiate the generic system dynamics with a multi-product inventory management scenario, illustrated in Fig. \ref{fig:sysarch}. As mentioned before, this is a part of the entire supply chain which was depicted in Fig. \ref{fig:sc-eg}. The goal of the controller is to maintain the inventory levels of all products above a minimum threshold, while simultaneously minimizing wastage due to products exceeding their use-by dates. The former objective is due to business requirements (which dictate that the shelves must not appear empty), while the latter objective is directly linked to incurred cost. The state $\bf{x}$ in (\ref{eq:mainsysdyn}) is given by the inventory levels of products. This is continuously depleted by sale of stock (the noise variable $\bf{w}$) as well as through spoiling of perishable inventory (the internal dynamics $F$). The inventory has to be periodically replenished (the control variable $\bf{u}$). The control actions are the quantities by which each product is replenished in each time step. The products have externally defined maximum inventory levels (shelf space), and their replenishment in each time step is constrained in terms of total weight and volume by the transportation capacity.

We define each element $x_i$ of state $\bf{x}$ to be the inventory level of product $i$, and the rate of order arrivals by the noise ${\bf{w}}\geq\bf{0}$. Inventory depletion due to spoiling of perishables is assumed to be at a fixed proportional rate $a_i$ for each product. The dynamics are thus $F({\bf{x}}(z)) = A\,{\bf{x}}(z)$, where matrix $A$ is a constant diagonal matrix with entries $-a_i$. Higher magnitudes of $a_i$ are for products that perish correspondingly faster. Since ${\bf{u}}(t)$ represents the replenishment actions at time $t$, its dimension is equal to that of $\bf{x}$ and $B$ is the identity matrix. Recall that we denote continuous time by $z$ and the discrete instants of replenishment by $t$. The dynamics are thus explicitly given by,
\begin{align}
\dot{\bf{x}}(z) & = \begin{pmatrix} -a_1 & \ldots & 0 \\  \vdots & -a_i & \vdots \\ 0 & \ldots & -a_p \\ \end{pmatrix} {\bf{x}}(z) -{\bf{w}}(z), \label{eq:scdyn} \\
{\bf{x}}(t)^+ & = {\bf{x}}(t)^- + {\bf{u}}(t). \nonumber
\end{align}

The inventory levels $x_i$ and control inputs $u_i$ are assumed to be continuous variables, which is accurate when the products are in liquid or gas form (for example, oil levels replenished periodically by tankers). In the case of unit-based products (for example, groceries), this assumption is approximately true as long as the maximum shelf capacity is significantly larger than the size of individual units. The inventory level of product $i$ between two time periods can be propagated by integrating relation (\ref{eq:scdyn}), which depends on the noise ${\bf{w}}(z)$. Note that the diagonal form of $A$ implies that the inventory equations can be solved independently between replenishment instants. We derive an expression for the change in inventory of product $i$ by first assuming that the integral of the noise (total orders) in one time step is, $W_i(t-1) =  \int_{t-1}^t w_i(z) \mathrm{d}z$.

Since the time period is assumed to last one unit, the average rate of orders within the time period is also equal to $W_i(t-1)$, which we shorten to $W_i$ for notational simplicity. Assuming that the rate $W_i$ is a constant in a given time period, the rate of depletion for product $i$ is given by,
\begin{equation*}
    \frac{\mathrm{d}x_i}{\mathrm{d}t} = -a_i\, x_i - W_i.
\end{equation*}

Integrating this differential equation from $(t-1)^+$ to $t^-$ (between two replenishment actions), we see that the inventory at the end of the time period is,
\begin{equation}
    x_i(t)^- = e^{-a_i} x_i(t-1)^+ - \frac{W_i}{a_i}\left( 1 - e^{-a_i}\right). \label{eq:stateupdate}
\end{equation}
Since the inventory cannot be a negative value, we assume that orders for any products that have no inventory are rejected and therefore $w_i(z)=0 \; \forall z\in [z^*,t)$ if $x_i(z^*)=0$. We note that even the functional form of the relationship (\ref{eq:stateupdate}) is not assumed known to the controller. It must either model the state and control relationships, or use a model-free technique as described in the next section. However, the relation (\ref{eq:stateupdate}) defines the state update relationship for each product in the inventory, which we use in the environment (simulation) portion of the RL algorithm. Assuming that the inventory levels and control inputs are normalized to the range $[0,1]$, the set of applicable constraints is as follows.
\begin{align}
{\bf{0}} \leq {\bf{x}}(t) & \leq {\bf{1}} \label{eq:inv} \\
{\bf{0}} \leq {\bf{u}}(t) & \leq {\bf{1}} \label{eq:control} \\
{\bf{0}} \leq {\bf{x}}(t)^- + {\bf{u}}(t) & \leq {\bf{1}} \label{eq:shelf} \\
{\bf{v}}^T\,{\bf{u}}(t) & \leq v_\mathrm{max} \label{eq:truckvol} \\
{\bf{c}}^T\,{\bf{u}}(t) & \leq c_\mathrm{max} \label{eq:truckwgt} 
\end{align}
Here, constraints (\ref{eq:inv}) and (\ref{eq:control}) are related to the range of acceptable values of each product. Constraint (\ref{eq:shelf}) states that the level of inventory just after replenishment (${\bf{x}}(t)^+$ according to (\ref{eq:repldyn})) cannot exceed the maximum inventory level. Constraints (\ref{eq:truckvol}) and (\ref{eq:truckwgt}) set maximum values on the total volume $v_\mathrm{max}$ and weight $c_\mathrm{max}$ of products replenished at a single time step, mimicking transportation capacity limitations. The vectors $\bf{v}$ and $\bf{c}$ are constants corresponding to the unit volume and weight multipliers for each product.

Inventory management is a multi-objective optimisation problem, with costs relating to (i) the reduction of product inventory below some threshold $\bf{\tau}$, commonly known as out-of-stock, (ii) the quantity $q_{\mathrm{waste},i}(t)$ of products wasted (spoiled) during the time period ending at $t$, and (iii) a fairness penalty on the variation in inventory levels across the product range. We include all three objectives in the reward function for reinforcement learning, with the fairness penalty represented by a term that penalizes the difference in inventory levels between the 95$^\mathrm{th}$ and 5$^\mathrm{th}$ percentile (denoted by $\Delta {\bf{x}}(t)^\mathrm{.95}_{\mathrm{.05}}$) across all products. For simplicity, we directly define the aggregate reward as,
\begin{equation}
R(t) = 1 - \underbrace{\frac{p_\mathrm{empty}(t)}{p}}_{\text{Out of stock}} - \underbrace{\frac{\sum_{i} q_{\mathrm{waste},i}(t)}{p}}_{\text{Wastage}} - \underbrace{\Delta {\bf{x}}(t)^\mathrm{.95}_{\mathrm{.05}}}_{\text{Percentile spread}},\label{eq:invreward}
\end{equation} 
where $p$ is the total number of products (size of $\bf{x}$), $p_\mathrm{empty}(t)$ is the number of products with $x_i\leq {\bf{\tau}}_i$ at the end of the period $[t-1,t)$. Since the maximum value of $p_\mathrm{empty}(t)$ is equal to $p$, the maximum value of $q_{\mathrm{waste},i}(t-1,t)$ is 1, and the maximum difference in inventory between two products is also 1, the theoretical range of the reward is $-2\leq R(t) \leq 1$. For practical purposes, the individual terms will be smaller than 1, and the vast majority of rewards should be in the range $[-1,1]$. The goal of the controller is to maximize the discounted sum of this reward as per (\ref{eq:reward}), at each time step $t$ given the dynamics (\ref{eq:scdyn}) and the constraints (\ref{eq:inv})-(\ref{eq:truckwgt}).

\section{Solution Methodology} \label{sec:method}

Before describing the data-driven training procedure for computing decisions, we note that the order rate ${\bf{w}}(z)$ plays a key role in the system dynamics (\ref{eq:scdyn})-(\ref{eq:stateupdate}). For simplicity, we define an estimator for the order rate $w_i$ of each product $i$ in the form of a trailing average of the order rates in the most recent $T$ time periods,
\begin{equation}
\hat{w}_i(z) = \frac{\int_{t-T}^{t} {w}_i(z') \mathrm{d}z'}{T},\forall \; z\in[t,t+1)\,i\in\{1,\ldots,p\}. \label{eq:forecast}
\end{equation}
There are several more sophisticated forecasting algorithms available in literature, but these are not the focus of this paper. Note that all the competing algorithms tested in Section \ref{sec:results} use the same values of forecast orders, in order to ensure a fair comparison of performance. Since we assume that each period lasts for a unit of time, the forecast for aggregate orders in one time period, $\hat{W}_i(t)$, is also given by (\ref{eq:forecast}).

We divide the RL approach into two logical steps: (i) definition of the state and action spaces and the rewards, and (ii) the specific RL architecture to be used. The primary difficulty in applying reinforcement learning to the problem at hand is the large scale of the problem. Depending on the number of products $p$, the state vector $\bf{x}$ and the decision vector $\bf{u}$ can both be of high dimension, and each element is continuous on the interval $[0,1]$. Further, the input should include context variables in addition to the current inventory level. We describe an RL algorithm for parallelised computation of replenishment decisions, by cloning the parameters of the same RL agent for each product and computing each element of the vector ${\bf{u}}(t)$ independently.

The advantage of this approach is that it splits the original problem into constant-scale sub-problems. Therefore, the same algorithm can be applied to instances where there are a very large (or even variable) number of products. This is a realistic situation in dynamic environments; for example, retailers are continuously adding new products and removing old products from their portfolio. However, the parallelised approach poses a challenge to the concurrent and fair optimisation of inventories across the product range. It is addressed by the reward and state definitions described below.

\textbf{Rewards: }
The key challenges with computation of individual elements of $\bf{u}$ are (i) ensuring that the system-level constraints (\ref{eq:truckvol})-(\ref{eq:truckwgt}) are met, and (ii) that all products are treated fairly. Both challenges are partially addressed using the reward structure. The fairness issue is addressed using the percentile spread term in (\ref{eq:invreward}), since it penalises the agent if some products have low inventories while others are at high levels. The volume and weight constraints are introduced as soft penalties in the following reward definition, adapted for individual decision-making.
\begin{align}
R_i(t) = & 1 - b_{i,\mathrm{empty}}(t) - q_{\mathrm{waste},i}(t)  \nonumber \\
& - \Delta {\bf{x}}(t)^\mathrm{.95}_{\mathrm{.05}} - \alpha\, \max (\rho-1,0), \label{eq:itemreward}
\end{align}
where $b_{i,\mathrm{empty}}(t)$ is a binary variable indicating whether inventory $i$ dropped below $\tau_i$ in the current time period, $\alpha$ is a constant parameter, and $\rho$ is the ratio of total volume or weight requested by the RL agent to the available capacity. We formally define this as,
\begin{equation*}
    \rho = \max \left( \frac{{\bf{v}}^T\,{\bf{u}}(t)}{v_\mathrm{max}},\; \frac{{\bf{c}}^T\,{\bf{u}}(t)}{c_\mathrm{max}} \right).
\end{equation*}
Equation (\ref{eq:itemreward}) defines the reward that is actually returned to the RL agent, as opposed to the true system reward defined in (\ref{eq:invreward}). If the aggregate actions output by the agent (across all products) do not exceed the available capacity ($\rho \leq 1$), then the average value of (\ref{eq:itemreward}) is equal to (\ref{eq:invreward}). This implies that maximising $R_i(t)$ is equivalent to maximising $R(t)$, as long as system constraints are not violated. The last two terms of (\ref{eq:itemreward}) are common to all products at a given time step $t$.

\begin{table}[h]
\caption{Features for each product $i$, used in the RL framework.}
\label{tab:ewrl}
\begin{center}
\begin{tabular}{|C{1.5cm}|l|}
\hline
Notation & Explanation \\
\hline
$x_i(t)$ & Current inventory level \\
$\hat{W}_i(t)$ & Forecast aggregate orders in $[t,t+1)$ \\
$\sigma_i$ & Historical std. dev. in forecast errors for $i$ \\
$v_i$ & Unit volume \\
$c_i$ & Unit weight \\
$l_i$ & Shelf life \\
${\bf{v}}^T\,{\hat{\bf{W}}}(t)$ & Total volume of forecast for all products \\
${\bf{c}}^T\,{\hat{\bf{W}}}(t)$ & Total weight of forecast for all products \\
\hline
\end{tabular}
\end{center}
\end{table}

\textbf{States and actions: }
Table \ref{tab:ewrl} lists the features used for the element-wise control computation. The first two features relate to the instantaneous state of the system with respect to product $i$. The next four inputs are product meta-data, relating to either long-term or constant behaviour. The quantity $\sigma_i$ is the standard deviation of forecast errors for that product, computed using historical data. Similarly, the shelf life $l_i$ is the normalized inverse of the \textit{average inventory loss} for product $i$ (the decrease in inventory not accounted for by orders, in a given time period). This is also computed empirically, and acts as an implicit estimator for the dynamics $A$. The meta-data distinguish between different product characteristics when they are processed sequentially by the same RL agent, by mapping individual products to the same feature space. The last two features in Table \ref{tab:ewrl} are derived features that provide indications of total demand on the system, with respect to the various constraints. These indicators act as inhibitors to the control action for product $i$, if the total demand on the system is high. They also help the agent correlate the last term in the observed rewards (the capacity exceedance penalty) with the state inputs. The output of the RL agent is ${u}_{i}(t)$, which is the desired action for product $i$ at time $t$. Individual actions are concatenated to form the control vector ${\bf{u}}(t)$.


\textbf{Neural network architecture: }
The computation of ${\bf{u}}_{i}(t)$ is carried out using a modified version of advantage actor critic (A2C) \cite{konda2000actor}, an algorithm that combines desirable attributes from both value and policy based approaches in RL. Two separate neural networks are used; one called the \textit{critic} and the other the \textit{actor}. The critic network is designed to evaluate the `goodness' of the current state of the system, while the actor is designed to select the optimal action (control decision) given the current state of the system. There are more sophisticated architectures based on this idea, including DDPG \cite{ddpg_modified}, TRPO \cite{schulman2015trust}, and PPO \cite{schulman2017proximal}. However, all of these in their default form have difficulty scaling to the large, continuous action spaces that we are interested in. We use DDPG as one of the algorithms used for comparison in Section \ref{sec:results}, where this becomes evident. On the other hand, the A2C approach has the advantage of faster (off-policy) learning, and ease of modification as described below.

In this work, the critic network accepts the 8 features as given in Table \ref{tab:ewrl}, and produces a scalar output which corresponds to the expected long-term reward, starting from the current state. Between the 8-neuron input layer and the 1-neuron output layer of the network, is a single hidden layer with 4 neurons. All neurons have \textit{tanh} activation. The network is trained using the \textit{keras} library in Python 3.5, using stochastic gradient descent with a learning rate of $0.025$, momentum of $0.8$, a batch size of $32$ samples, and mean squared loss with respect to the TD(0) error (standard methodology for A2C). The actor network processes the same 8 inputs as the critic, but its output is a probability distribution over a user-defined set of quantized actions between 0 and 1. In general, the output layer could be any set of $n$ quantized action values. Because of the larger output size, the actor network has two hidden layers with $2n$ neurons each. The hidden layers have \textit{tanh} activation while the output layer has \textit{relu} activation. Its training methodology is different from the basic A2C, and is described below.

\textbf{Modified training approach: }
The TD(0) discounted reward is used to compute the advantage $\delta_i$ for product $i$ with respect to the value estimated by the critic. Intuitively, a positive $\delta_i$ should encourage the actor to increase the probability of choosing the same action again. However, a subtle difference exists between the outputs of a typical discrete choice problem (such as classification) and the current problem. In our case, the outputs map to specific scalar values, and neighbouring outputs correspond to \textit{similar} actions. Therefore, a loss function such as cross-entropy (which has a hard distinction between \textit{right} and \textit{wrong} outputs) is not used here, and we also forgo the usual policy-gradient based training methodology for the actor. Instead, we use a mean square loss with respect to an adjusted probability distribution as follows.

Let us assume that the chosen output for a product $i$ is $j$ with an activation $a(j)$, and the realized actor delta from the TD(0) trace is $\delta_i$. Then the target value of each action $j^*\in\{1,\ldots,n\}$ is set to $\left(a(j^*)+\frac{\delta_i}{q\,(|j-j^*|+1)}\right)$. The target vector is then re-normalized to sum up to 1, and the actor is trained using a mean square loss compared to this updated and smoothed vector. The action corresponding to each product in each time period is stored as a unique sample in an experience buffer, with batch training after every 32 time periods of $32p$ samples. The combined vector of desired control actions is denoted by $\bf{u}$. Since the output actions are rectified linear units (\textit{relu}), these actions need not satisfy constraints (\ref{eq:control})-(\ref{eq:truckwgt}). Therefore, a constrained version of the control action is calculated in two steps. First, ${\bf{u}}_i(t)$ is clipped to a maximum of $(1-{\bf{x}}_i(t))$, in order to satisfy (\ref{eq:control}) and (\ref{eq:shelf}). Second, constraints (\ref{eq:truckvol}) and (\ref{eq:truckwgt}) are enforced by proportionally reducing the desired quantities,
\begin{equation}
{\bf{u}}_{con}(t) = {\bf{u}}(t)\cdot \min \left( 1,\; \frac{v_\mathrm{max}}{{\bf{v}}^T\,{\bf{u}}(t)},\; \frac{c_\mathrm{max}}{{\bf{c}}^T\,{\bf{u}}(t)} \right). \label{eq:constraint}
\end{equation}
The capacity exceedance penalty according to $\rho$ ensures that this reduction to feasible values is part of the training rewards for the RL algorithm, and is rarely required after training.

\begin{figure}[h]
    \centering
    \includegraphics[width=0.65\textwidth]{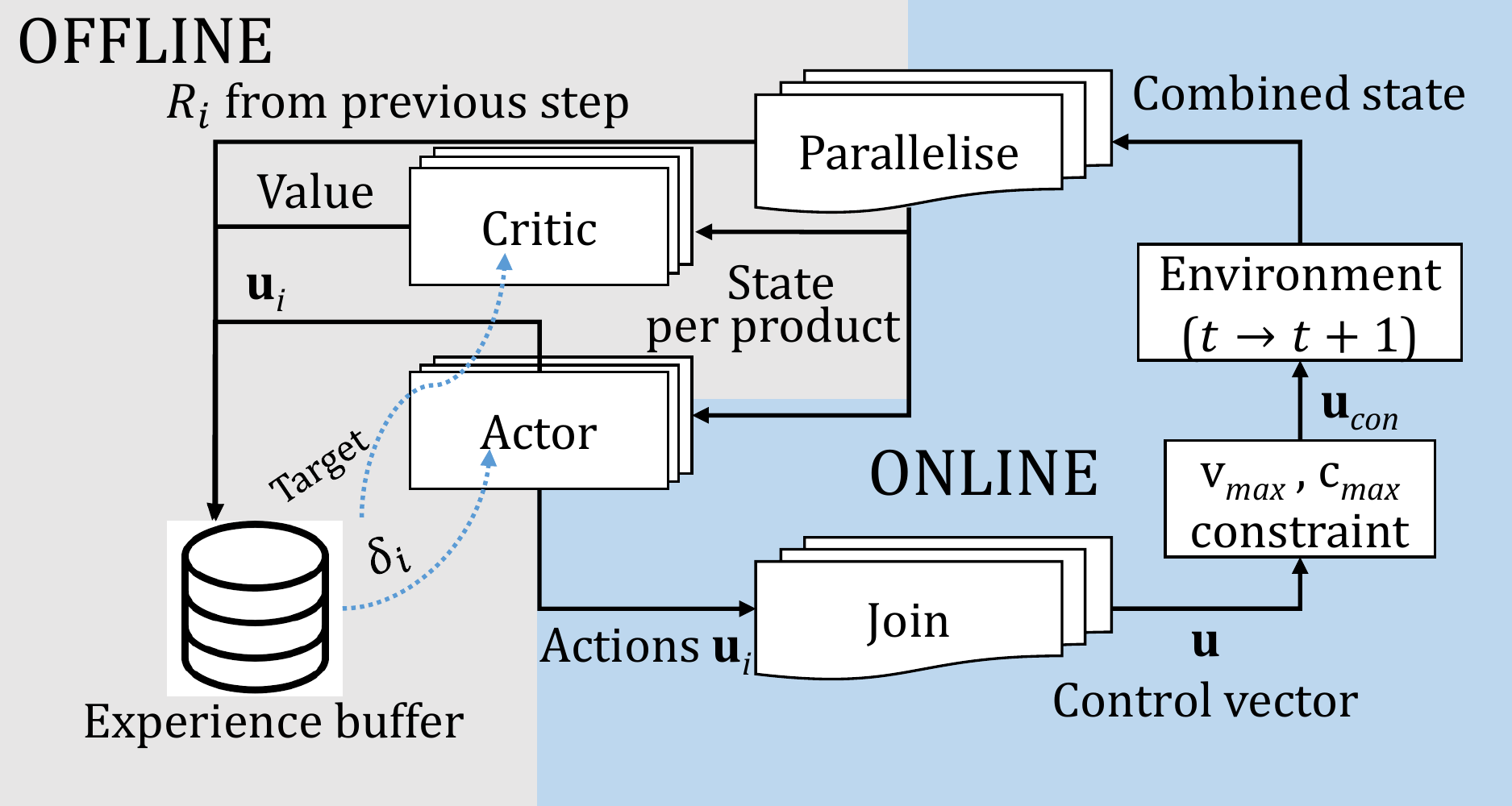}
    \caption{Architecture of the modified A2C algorithm, with parallelised online decision-making. The workflow in the left half (gray background) of the figure operates offline, for training. Online decisions require only the right half (blue background).}
    \label{fig:AC1}
\end{figure}

\section{Experiments and results} \label{sec:results}

In this section, we describe a set of experiments run to test the algorithm performance in comparison with prior algorithms in literature. The data used for experiments is based on a public source, and is described further below.

\subsection{Data for experiments}

We use a public data set for brick and mortar stores \cite{Kaggle} as the basis for the experimentation. The original data set includes purchase data for 50,000 product types and 60,000 unique customers. However, it does not contain meta-data about the products (dimensions, weight) and also does not specify date of purchase (although it specifies time of day and the day of week). Instead, it measures the number of days elapsed between successive purchases by each customer. We obtain data in the format required by the current work, by assigning a random date to the first order of each unique customer while respecting the day of week given in the original data set. This implicitly assigns specific date and time stamps to each purchase. Additionally, we assign dimension and weight to each product type based on the product label (which is available in the original data set). Since this is a manual process, we only use a subset of 220 products from the full data set, corresponding to one full grocery aisle in the store. 

\begin{figure}[h]
    \centering
    \includegraphics[width=0.65\textwidth]{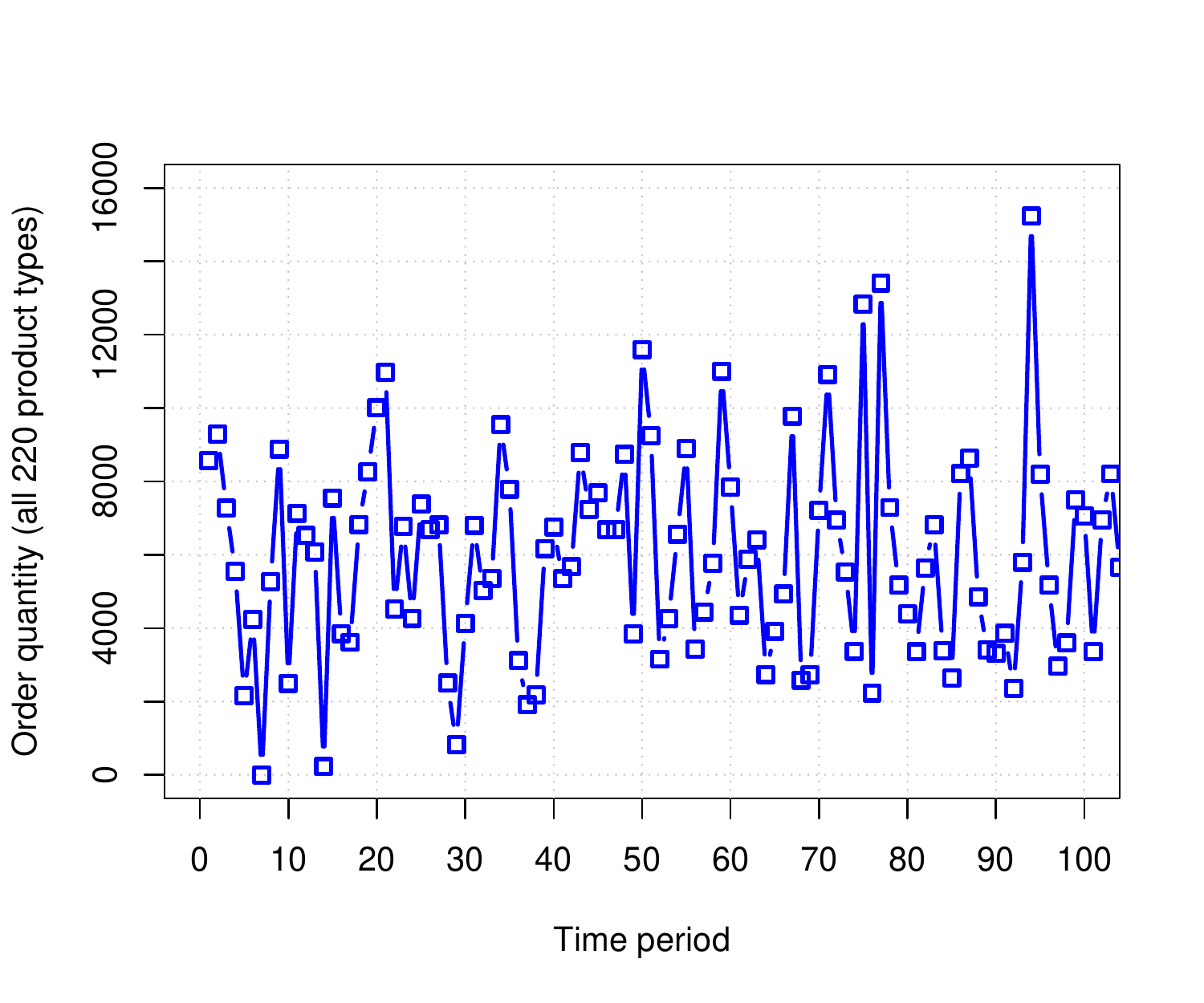}
    \caption{Variation in order quantities (summed over all product types) for a sample of the training data set.}
    \label{fig:orderdata}
\end{figure}

The data set thus formed spans a period of 349 days. We assume that stock replenishment happens once every 6 hours (4 times a day), resulting in 1396 time periods. Of these, we use the first 900 time periods for training, and use the final 496 time periods for testing. A sample of variation in order quantities over time is shown in Fig. \ref{fig:orderdata}. We observe a fair degree of variation from one time period to another, including outliers on both sides of the bulk of data. The semi-synthetic data set with 220 product types is used for comparison against other approaches in Section \ref{sec:results}. Additionally, we perform the same data manipulation on an independent set of 100 products, and report the training and test results for our algorithm. In both sets of data, the volume and weight constraints ($v_\mathrm{max}$ and $c_\mathrm{max}$) are set slightly lower than the average volume and weight of orders, ensuring that constraints (\ref{eq:truckvol})-(\ref{eq:truckwgt}) are active.


\subsection{Baseline algorithms}

We use a number of methodologies from literature, to benchmark the performance of the proposed approach (which we call A2C\_mod). For reasons explained in Section \ref{sec:related}, we do not use approximate DP or imitation learning. Instead, we use a heuristic from a previous paper and considered to be standard in supply chain, and also several RL algorithms.

\textbf{Heuristic based on proportional control: }
We use a modified version of s-policy \cite{nahmias1994optimizing}, a standard heuristic in literature which aims to maintain inventory at some predefined constant level. Since the control matrix $B$ is assumed to be known and is equal to the identity matrix according to (\ref{eq:scdyn}), the simplest replenishment approach is to model the system as a regulator with a reference state signal, and to use a linear proportional controller to compute the actions \cite{ogata2002modern}. The reward structure (\ref{eq:invreward}) indicates that low inventory (state) levels carry the risk of stock-out (zero inventory), while high inventory levels will lead to high wastage. Therefore, it is probable that holding inventory levels at some intermediate value between $0$ and $1$ will lead to good rewards. If we define $\bf{x}^*$ to be the \textit{target} level of inventories for the products, the reference signal is the sum of the target level and the expected inventory depletion due to orders in the next time period. The canonical proportional control relationship is the product of the state error (${\bf{x}}^* + \hat{\bf{W}}(t) - {\bf{x}}(t)^-$) and a gain matrix $K$. We note that the control matrix $B$ is the identity matrix, and therefore does not appear. Furthermore, the control is restricted to non-negative values, and therefore the desired replenishment quantity is given by,
\begin{equation}
{\bf{u}}_{pr}(t) = \max\left[ 0, {\bf{x}}^* + \hat{\bf{W}}(t) - {\bf{x}}(t)^-\right]. \label{eq:he_desired}
\end{equation}
The desired action ${\bf{u}}_{pr}(t)$ according to (\ref{eq:he_desired}) satisfies constraints (\ref{eq:control}) and (\ref{eq:shelf}), since ${\bf{0}}\leq {\bf{x}}^* \leq {\bf{1}}$. However, it is not guaranteed to satisfy the total volume and weight constraints (\ref{eq:truckvol}) and (\ref{eq:truckwgt}). Therefore, the final control vector is computed by proportional reduction analogous to (\ref{eq:constraint}).

\textbf{Other RL approaches: }
We use three types of related RL approaches in addition to the modified A2C, for performance comparison. In order to ensure fairness, we retain the same input and output schema and forecast values for all these algorithms. The closest related approach is to use vanilla A2C with categorical cross-entropy loss. Second, we implement a DQN approach (with separate target and online networks) by dropping the actor network from the A2C and modifying the critic network to output value estimates for each of the $n$ quantised actions. Third, we implement DDPG \cite{ddpg_modified} which is based on the actor-critic framework and has continuous action output. We provide the same state as provided to the A2C method to the actor, which outputs replenishment quantities for each product. The critic accepts the state vector as well as the replenishment quantity, and outputs action-value estimate for each product. The weights of actor and critic are shared between all products for fairness. During training, product indices are randomly shuffled to negate any ordering biases.

\subsection{Results}

\begin{figure}[h]
    \centering
    \includegraphics[width=0.65\textwidth]{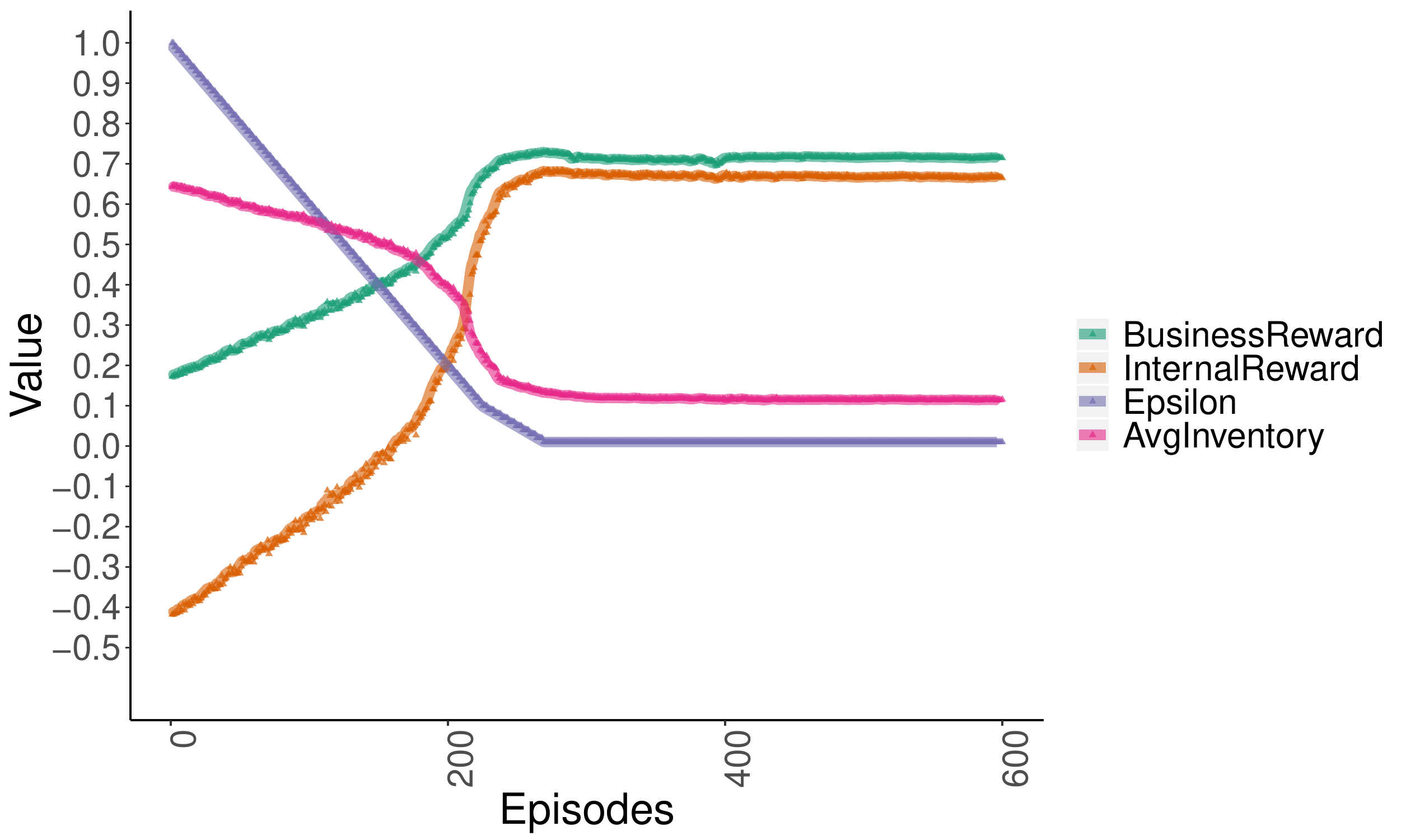}
    \caption{Components of A2C\_mod during training on 220 products.}
    \label{fig:A2CResults}
    \vskip-10pt
\end{figure}

We ran each algorithm for 600 training episodes, each containing the 900 time steps in training data. Various initial inventory levels were tried, and the results were largely invariant (the effect of initial state quickly diminishes in a 900-step episode). The training progress of A2C\_mod is shown in Fig. \ref{fig:A2CResults}. Of key interest are the curves for `business reward', which is the overall system reward based on (\ref{eq:invreward}), and the `internal reward', which is the average per product as per (\ref{eq:itemreward}). As training progresses, the algorithm learns to minimise the difference between the two curves, by minimising capacity overshoot (based on $\rho$). It also learns to reduce the average inventory levels so as to minimise wastage, while keeping them above the stock-out level $\bf{\theta}$. Nearly identical behaviour is observed when the same algorithm (without change in architecture or hyperparameters) is trained on an independent set of 100 products. Only the truck volume and weight capacities ($v_\mathrm{max}$ and $c_\mathrm{max}$) are changed in this run, in order to reflect the different meta-data and order quantities.

\begin{figure}[h]
    \centering
    \includegraphics[width=0.65\textwidth]{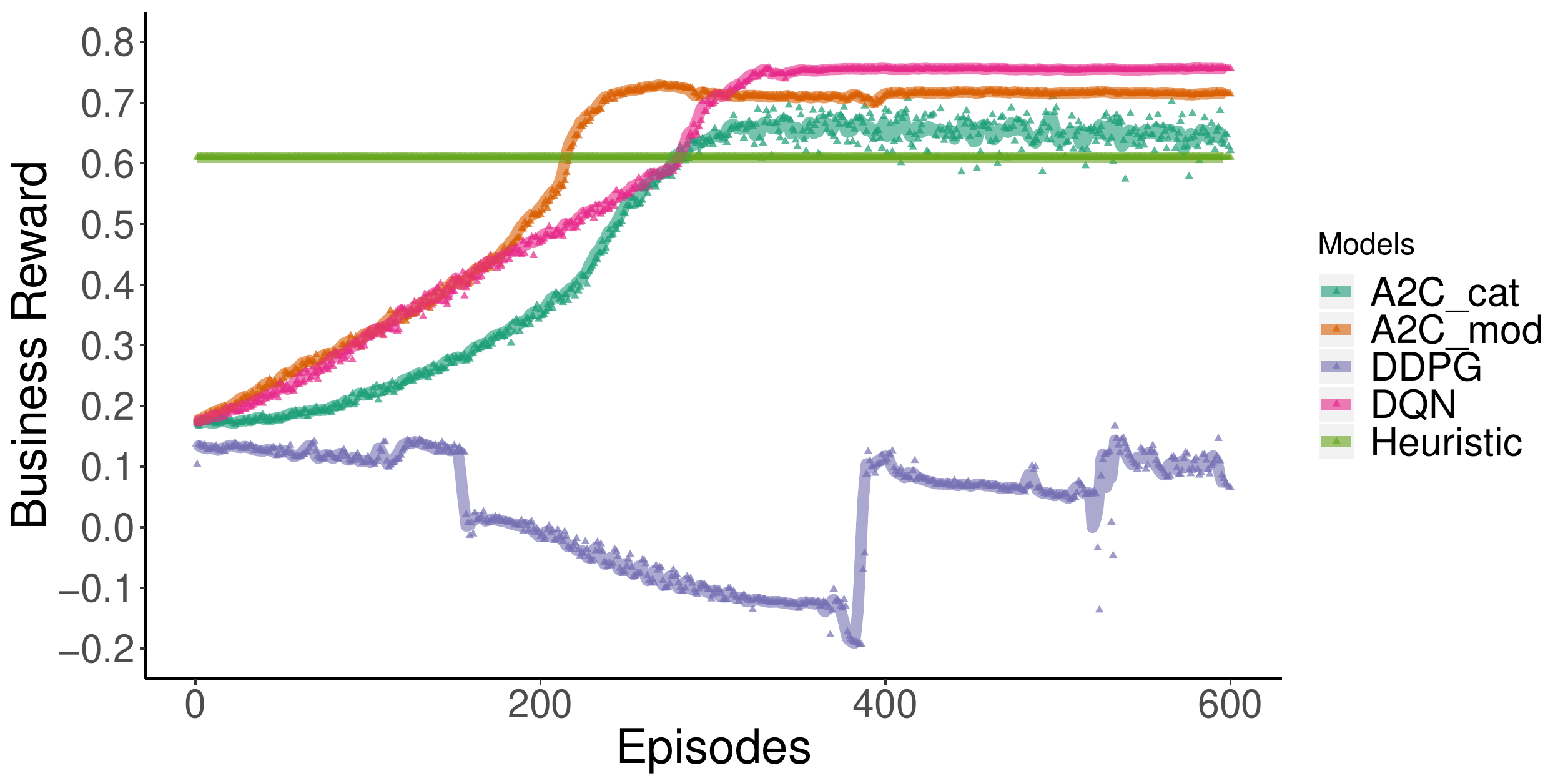}
    \caption{Rewards during training for all competing algorithms.}
    \label{fig:Results}
    \vskip-10pt
\end{figure}

The average rewards as per (\ref{eq:invreward}) for all competing algorithms over the course of training on the 220-product data set, are shown in Fig \ref{fig:Results}. The results on the independent test data set are compiled in Table \ref{tab:results}. The proposed algorithm (A2C\_mod) has the fastest growth in reward and the second-best performance after 600 episodes. Among the baseline algorithms, the proportional control heuristic has no training, and its performance is thus flat. A2C\_cat differs only in the loss function from A2C\_mod, but the performance gap is persistent. It also retains a high degree of variance towards the end of training. DDPG converges to the lowest reward on both training and test data in 600 episodes. DQN learns at a slower rate compared to A2C\_mod, but converges to a higher reward. Possible reasons are as follows.

\begin{table}[h]
\caption{Performance on test data set.}
\label{tab:results}
\begin{center}
\begin{tabular}{lll}
\toprule
     Model & Training &   Testing \\
\midrule
   A2C\_mod &    0.709 &  0.723 \\
   A2C\_cat &    0.671 &  0.718 \\
       DQN &    0.757 &  0.737 \\
      DDPG (7500 episodes) &    0.692 &   0.702 \\
 Heuristic &     0.63 &  0.607 \\
\bottomrule
\end{tabular}
\end{center}
\end{table}

\begin{figure}[h]
    \centering
    \includegraphics[width=0.65\textwidth]{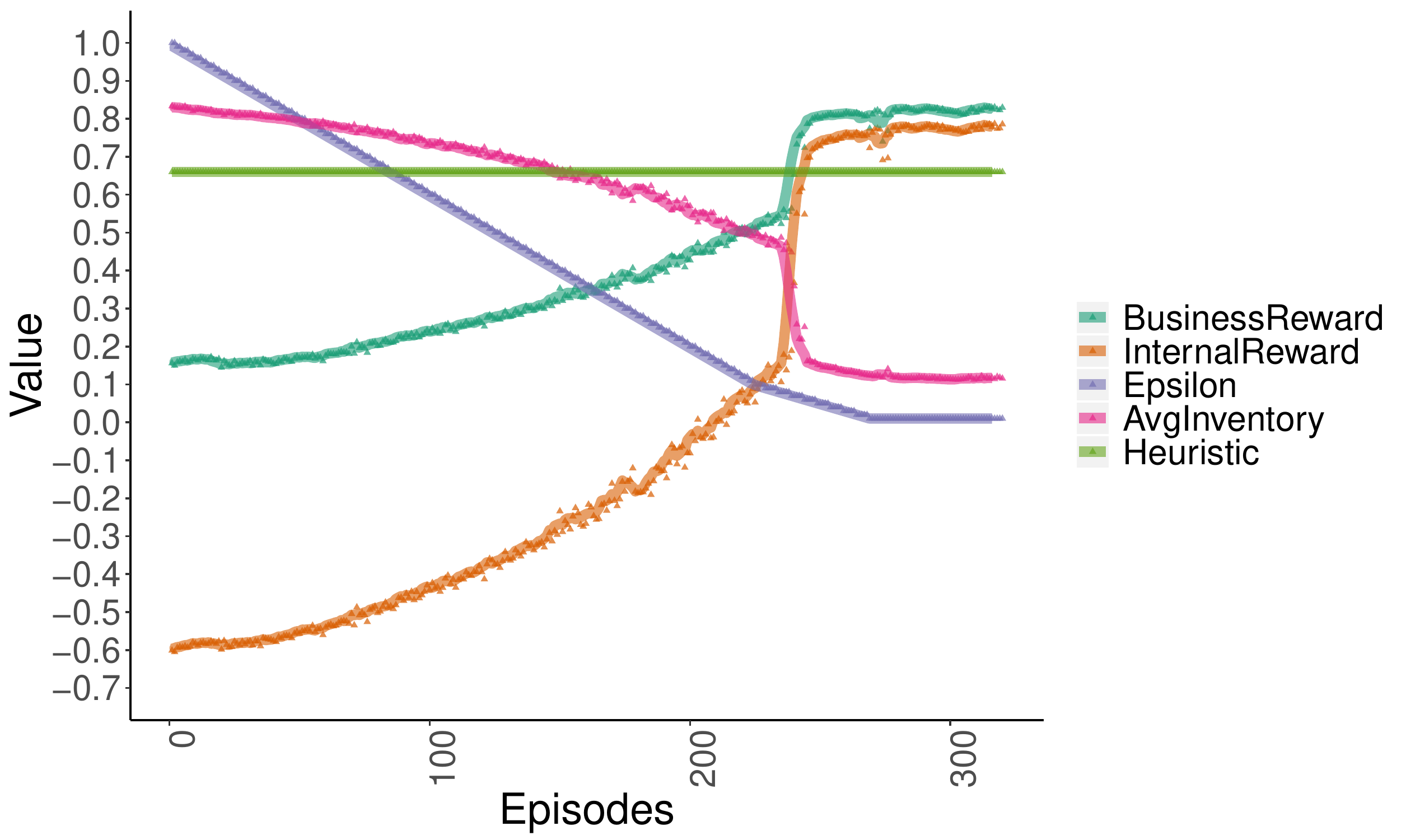}
    \caption{Components of A2C\_mod during training on 100 products.}
    \label{fig:A2CResults_100}
\end{figure}

\begin{figure}[h]
    \centering
    \includegraphics[width=0.65\textwidth]{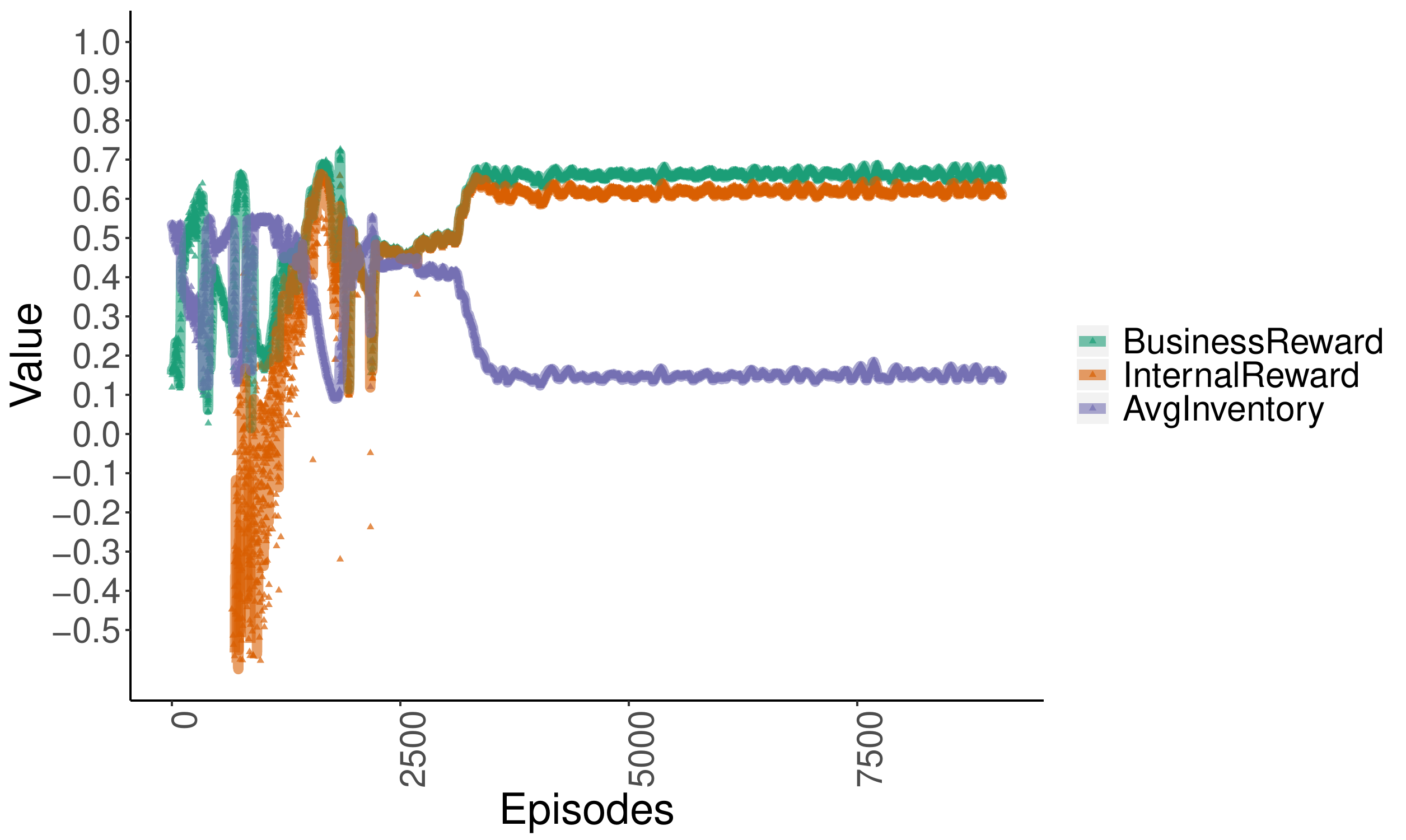}
    \caption{DDPG with longer training.}
    \label{fig:DDPG_Results}
\end{figure}

\begin{figure}[h]
    \centering
    \includegraphics[width=0.99\textwidth]{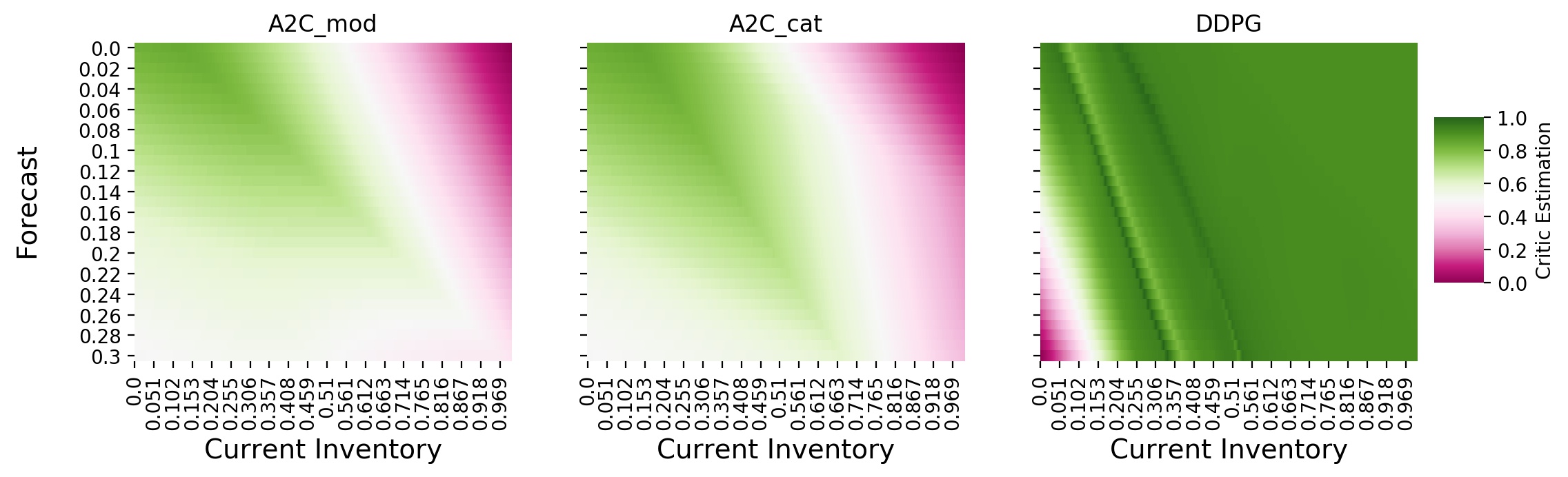}
    \vskip-5pt
    \caption{Critic estimate as a function of inventory and forecast.}
    \label{fig:Value}
\end{figure}

First, the large variations in DDPG are due to its continuous action output and hence it requires longer training. As shown in Fig. \ref{fig:DDPG_Results}, DDPG also converges to a similar reward (0.69) if trained for over 3000 episodes (5 times the length of training in Fig. \ref{fig:Results}). Second, we refer to the critic outputs in Fig. \ref{fig:Value}. Note that DQN is not included in Fig. \ref{fig:Value} since there is no critic in this method, and that the DDPG results are from the fully trained version after 7500 episodes. The two A2C-based methods have broadly similar value estimates, with the highest values seen for low forecast and moderately low inventory levels. The worst values for A2C are near the top right, where forecast is low while inventory is high. There is no action available to address this problem, which leads to high wastage. The lowest values for DDPG are at the low-inventory high-forecast corner (bottom left). This is also an undesirable state, but it can be fixed by a large replenishment action. Note that the DDPG value estimate is based on both state and action, while the A2C values are based only on state.

\begin{figure}[h]
    \centering
    \includegraphics[width=0.99\textwidth]{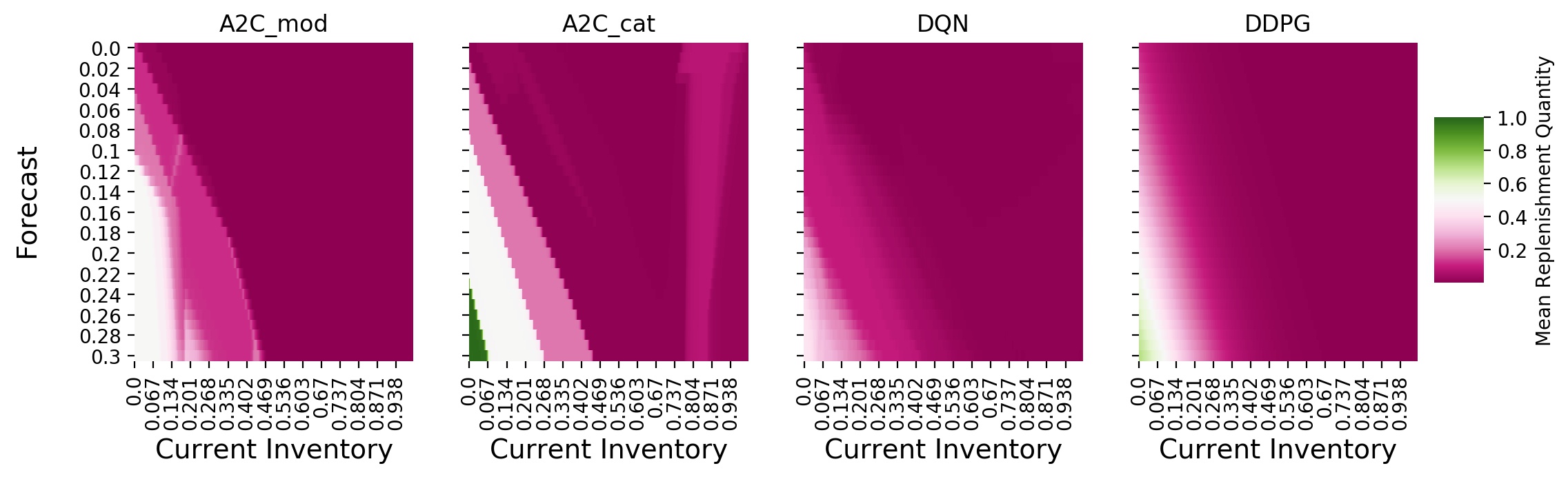}
    \vskip-5pt
    \caption{Mean replenishment quantity across products as a function of inventory and forecast.}
    \label{fig:policy_heatmap}
\end{figure}

Finally, in Fig. \ref{fig:policy_heatmap} we compare the requested replenishment action as a function of two features (current inventory and forecast), averaging over all other feature values. We note that A2C\_cat has sharp edges in the policy, indicating that the actions are based primarily on inventory and forecast. The other three approaches have broadly similar policies, with DQN having the smoothest variation. This could be the reason for its mild advantage over all other methods in training as well as test data. The key takeaway from the results is that while each RL approach has its advantages and disadvantages (stability, ease of training), the individual decision-making framework (product-by-product action computation) with modified training scheme (mean squared error loss smoothed over all actions, used in A2C\_mod and DQN) is able to yield high solution quality with low computational cost, and is also able to learn to stay within the aggregate system capacity constraints.


\section{Conclusion}

This paper proposed the use of reinforcement learning for multi-objective optimisation of online decision-making in dynamic systems. We showed that systems described by vector differential equations could be formulated as RL control problems, and instantiated this with the example of multi-product inventory management. We modified the advantage actor critic (A2C) algorithm by noting the close relationship between actions in a quantised space, and showed that the resulting method could outperform other approaches on at least one problem instance. In future work, we wish to identify standard problems and data sets to more thoroughly test our hypotheses.

\section*{Acknowledgment}

The authors would like to thank Dheeraj Shah and Padmakumar Ma from TCS Retail Strategic Initiatives team, for their help in defining the inventory management use case.

\bibliographystyle{unsrt}
\bibliography{ieee-tac-rl}


\end{document}